\documentclass[10pt,twocolumn,letterpaper]{article}
\usepackage[pagenumbers]{iccv}
\usepackage{booktabs}
\usepackage{multirow}
\usepackage{makecell}

\definecolor{iccvblue}{rgb}{0.21,0.49,0.74}
\usepackage[pagebackref,breaklinks,colorlinks,allcolors=iccvblue]{hyperref}

\newif\ifappendix
\appendixfalse %

\newcommand{\enableappendix}{\appendixtrue}

\newcommand{\acref}[1]{%
  \ifappendix
    \cref{#1}%
  \else
    the supplementary material%
  \fi
}

\newcommand{\mname}{VMem}%
\newcommand{\mnamel}{Surfel-Indexed View Memory}%

\makeatletter
\renewcommand{\paragraph}{%
  \@startsection{paragraph}{4}%
  {\z@}{-0.5em}{-0.5em}%
  {\normalfont\normalsize\bfseries}%
}
\makeatother

\def\paperID{3609}
\def\confName{ICCV}
\def\confYear{2025}

\title{%
\mname: Consistent Interactive Video Scene Generation with Surfel-Indexed View Memory%
}
\author{%
Runjia Li%
\quad%
Philip Torr%
\quad%
Andrea Vedaldi%
\quad%
Tomas Jakab \\[0.4em]
University of Oxford \\[0.1em]
\small \href{https://v-mem.github.io}{\texttt{v-mem.github.io}}%
}

\enableappendix %

\begin{document}

\twocolumn[\maketitle\vspace{-2em} \begin{center}
    \includegraphics[width=\textwidth]{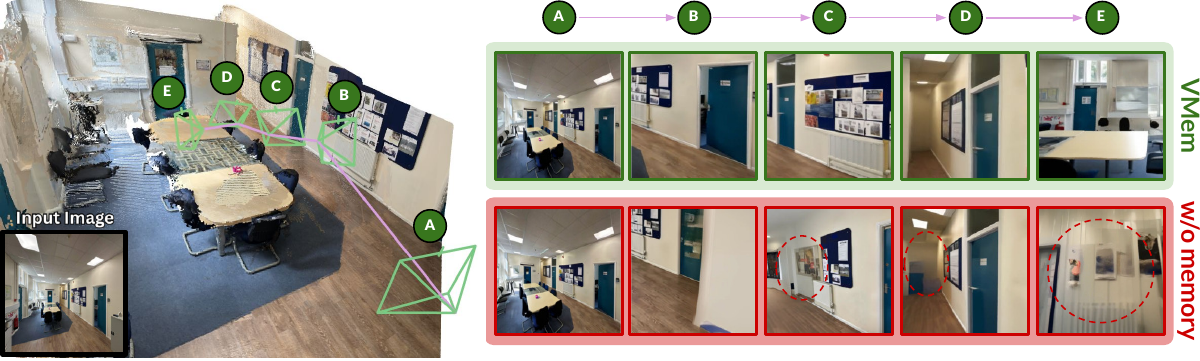}
\end{center}
\vspace{-5mm}
\captionof{figure}{\mname~enables autoregressive scene generation from a single image along user-defined trajectories.
The \textcolor{OliveGreen}{green} region shows results with the proposed memory module, maintaining coherence when generating previously seen parts of the scene.
The \textcolor{BrickRed}{red} region, without memory, exhibits degradation highlighted with red ellipses, demonstrating \mname~is effective for consistent scene generation.}
\bigbreak]

\begin{abstract}
We propose a novel memory module for building video generators capable of interactively exploring environments.
Previous approaches have achieved similar results either by out-painting 2D views of a scene while incrementally reconstructing its 3D geometry---which quickly accumulates errors---or by using video generators with a short context window, which struggle to maintain scene coherence over the long term.
To address these limitations, we introduce \textit{\mnamel} (\textit{\mname}), a memory module that remembers past views by indexing them geometrically based on the 3D surface elements (surfels) they have observed.
\mname{} enables efficient retrieval of the most relevant past views when generating new ones.
By focusing only on these relevant views, our method produces consistent explorations of imagined environments at a fraction of the computational cost required to use all past views as context.
We evaluate our approach on challenging long-term scene synthesis benchmarks and demonstrate superior performance compared to existing methods in maintaining scene coherence and camera control.
\end{abstract}

\section{Introduction}%
\label{sec:intro}

We consider the problem of generating long videos that explore an imagined space following a camera path specified interactively by the user.
In this paradigm, the user tells the model which camera path to follow for the next few frames, observes the generated content, and then decides where to explore next based on what they have seen.
For example, exploring a house may involve visiting the kitchen, the living room, and the bathroom, eventually returning to the kitchen.
Throughout the video, the scene must remain consistent, ensuring that the kitchen looks the same upon return.
Generating such videos is essential for immersive applications such as games, where players can navigate generated worlds.
However, even recent large-scale interactive video generators such as Google's Genie~2~\cite{parker-holder24genie} struggle to achieve this goal.

This problem has so far been addressed by two types of methods.
First, \emph{outpainting-based} methods~\cite{wiles2020synsin, SceneScape, hoellein2023text2room, muller2024multidiff, popov24camctrl3d, yu2024viewcrafter, seo2024genwarp, koh2023simple, liu2021infinite, rockwell2021pixelsynth} iterate between generating new 2D views of the scene and estimating its 3D geometry.
They use the estimated geometry to partially render a new viewpoint and then employ an outpainting model to complete the missing parts, thus adding one more image to the collection.
However, errors in outpainting, 3D reconstruction, and stitching accumulate over time, leading to severe degradation of the generated content after a short while.

Second, \emph{multi-view/video-based} methods~\cite{geogpt, photometricnvs, ren2022look, wang2024motionctrl, parker-holder24genie} condition novel view generation on previous views using a geometry-free approach that does not explicitly estimate the scene geometry.
While this avoids the accumulation of errors in the reconstructed 3D scene geometry, it comes at a high computational cost, limiting the number of conditioning views to a small context window of recent frames.
This constraint hurts the long-term consistency of the generated images.

\begin{figure*}[t!]
\centering
\includegraphics[width=\textwidth]{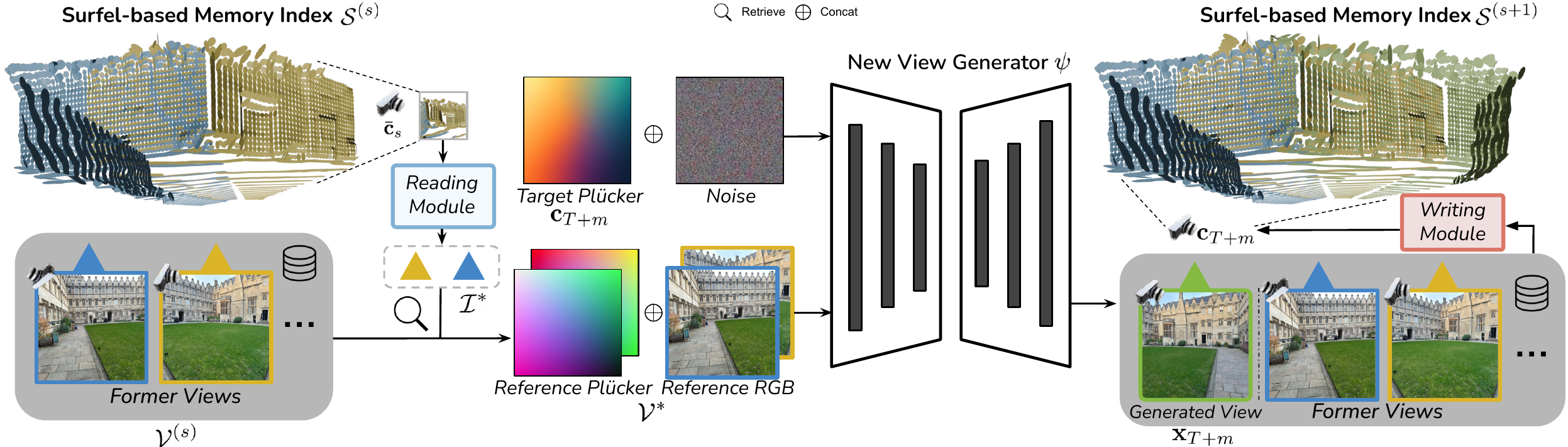}
\caption{\textbf{Method.}
Given target camera viewpoints $\{\mathbf{c}_{T+m}\}_{m=1}^M$, we query our \mnamel~to retrieve the most relevant $K$ past views $\mathcal{V}^* \subset \mathcal{V}^{(s)}$ where $\mathcal{V}^* = \{v_t\}_{t=1}^K$ as references.
Retrieved reference images $\mathbf{x}_t$ along with Plücker embeddings of both reference camera poses $\mathbf{c}_t$ and target camera poses $\{\mathbf{c}_{T+m}\}_{m=1}^M$ are fed into generator $\psi$ to synthesize novel views $\{\mathbf{x}_{T+m}\}_{m=1}^M$.
After generation, the surfel-indexed memory is updated $\mathcal{S}^{(s)} \rightarrow \mathcal{S}^{(s+1)}$ by appending new view indices $\{T+m\}_{m=1}^M$ to existing surfels or creating new surfels based on geometry of the generated views.
This is repeated autoregressively, enabling long-term consistent generation.
}
\label{fig:method}
\end{figure*}

In this work, we revisit the second class of methods and propose conditioning not on the most recently generated views, but on the \emph{most relevant ones} for generating the next image, thereby maintaining a high degree of consistency within a limited computational budget.
Given a novel viewpoint of the scene, the most relevant past views are those that have already observed the parts of the scene currently being generated.
This implies that, for each part of the scene, we must remember which views have previously observed it and retrieve them from memory.

To achieve this, we introduce \textit{\mnamel}, abbreviated as \textit{\mname}, a memory module that anchors previous views to the 3D surface elements they observe.
Given a new viewpoint, we retrieve the past views that best capture the currently observed surfaces and use them to condition novel view generation.
To create the memory, we estimate the geometry of each new view using an off-the-shelf point map predictor.
This is similar to outpainting methods but, crucially, we do not use the estimated geometry as the final representation of the scene;
instead, we use it to construct \mname{}, which is a memory of past views.
We represent the scene geometry using surfels, which are more robust compared to meshes and can represent occlusions compared to point clouds.
Each surfel stores in its attributes a set of indices corresponding to past viewpoints that observed it.

To retrieve relevant past viewpoints, we render the surfels with their attributes from the novel viewpoint and splat them onto an image grid.
Each pixel in the resulting image then corresponds to a set of viewpoint indices.
We select the top $K$ most frequently represented viewpoint indices and use them to retrieve past views from the database, where each view is represented by an RGB image and its corresponding camera parameters.
A key advantage of our approach over outpainting-based methods is that it does not require highly accurate scene geometry.
As long as we successfully retrieve the most relevant past views, our method remains robust.

By leveraging \textit{\mnamel}, we significantly reduce the memory and computational burden of conditioning on a large number of previous views, as required by multi-view/video-based methods, while improving long-term consistency across generated novel views.

Our approach represents a step toward scalable, realistic, and long-term autoregressive scene generation, making the following contributions:
\begin{enumerate}
    \item We introduce \mnamel, a \textit{plug-and-play} module to index past views geometrically and use them to condition novel view generation.
    \item We show that our method can generate long-term coherent videos of scenes and outperforms existing approaches.
    \item We demonstrate that \mnamel~achieves comparable performance with $4\times$ fewer context views, delivering a $12\times$ speedup.
    \item We validate our approach on challenging benchmarks, outperforming the current open-source state-of-the-art methods.
\end{enumerate}

\section{Related work}%
\label{sec:related}

Novel view synthesis (NVS) is a challenging and ill-posed problem that can be categorized into two main categories: view interpolation methods~\cite{niemeyer2022regnerf, chen2021mvsnerf, chen_single-stage_2023, chen2024mvsplat, chen2024mvsplat360, wu2024reconfusion, truong2023sparf, gao2024cat3d}, which generate views close to given inputs, and view extrapolation or autoregressive view generation, where novel views extend significantly beyond the original scene, introducing substantial new content.
The latter is particularly difficult from single images.
Between these lie single-view scene reconstruction methods, which succeed mainly for single-object scenes~\cite{melas-kyriazi23realfusion, shi23zero123:, hong24lrm:, szymanowicz24splatter, boss2024sf3d, xiang2024structured} or highly bounded scenes~\cite{szymanowicz2024flash3d}.
However, their extrapolation capabilities remain limited.
Most relevant to our work are single-image view extrapolation models, which fall into two categories: those incorporating explicit geometric modeling with inpainting, and those based directly on image or video generation.

\paragraph{Inpainting-based view extrapolation.}
Inpainting-based methods~\cite{wiles2020synsin, SceneScape, hoellein2023text2room, muller2024multidiff, popov24camctrl3d, yu2024viewcrafter, seo2024genwarp, koh2023simple, liu2021infinite, rockwell2021pixelsynth, yu23wonderjourney:, yu25wonderworld:} use pre-trained 3D reconstruction models to generate 3D representations—such as meshes, point clouds, or Gaussian splats—from images.
These representations are reprojected into novel views, where 2D inpainting fills missing regions.
SceneScape~\cite{SceneScape} reconstructs meshes from images using pre-trained depth estimators.
Diffusion-based inpainting then completes the projected novel view, which is reprojected back to refine and extend the mesh.
This process is iterated to generate novel views.
MultiDiff~\cite{muller2024multidiff} uses depth estimators to warp reference images into multiple novel views, training diffusion models to inpaint missing regions simultaneously.
CamCtrl3D~\cite{popov24camctrl3d} follows a similar approach but adds ray maps to condition diffusion models for view synthesis.
ViewCrafter~\cite{yu2024viewcrafter} uses pre-trained pointmap estimators to create point clouds and trains video diffusion models for inpainting.
GenWarp~\cite{seo2024genwarp} warps 2D coordinate representations of reference views into novel views and uses them to condition generation.

While these methods improve geometric consistency, they are susceptible to errors in depth or pointmap estimation, which propagate distortions across generated views.
Once 3D representations are constructed, inaccuracies from depth or pointmap estimation become difficult to correct.
Moreover, scaling to large scenes is computationally intensive, as storing and processing high-fidelity 3D representations requires substantial memory.

\begin{figure*}[h]
\centering
\includegraphics[width=0.95\textwidth,clip,trim=0 0.4cm 0 0cm]{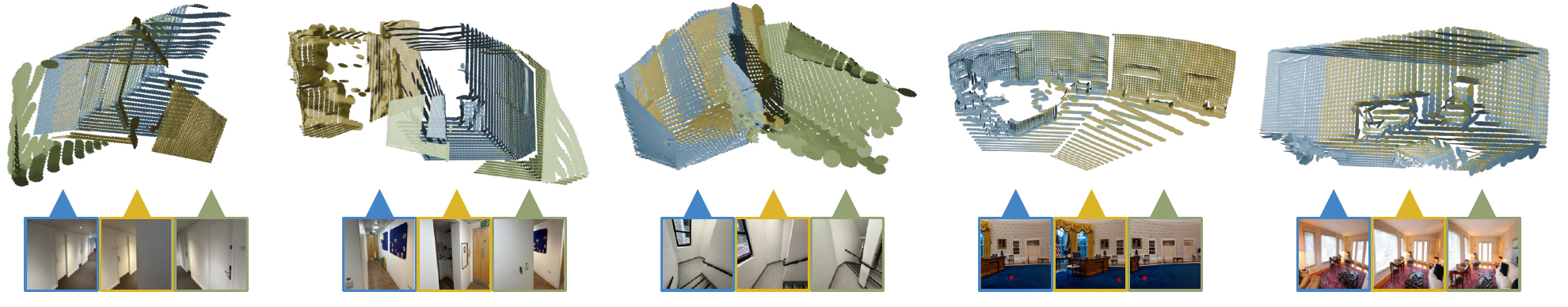}
\caption{\textbf{Surfel-based memory index.}
Each surfel stores indices of views that observed it.
We color-code each surfel by contributing view indices.
This spatial index enables retrieval of relevant past views: when generating a novel view, we identify visible surfels from the target viewpoint and retrieve views that previously observed those same regions, naturally accounting for occlusion.
}%
\label{fig:surfel_demo}
\end{figure*}

\paragraph{Multi-view-based view extrapolation.}
Another category~\cite{geogpt, ren2022look, tseng2023consistent, photometricnvs, wang2024motionctrl, gao2024cat3d, seva} avoids explicitly modeling 3D scene geometry.
Instead, these approaches generate novel views by conditioning on previously rendered or generated views.
GeoGPT~\cite{geogpt} uses autoregressive likelihood models to synthesize novel views from single images.
Building on this, LookOut~\cite{ren2022look} improves consistency by generating view sequences along predefined trajectories, conditioning on up to two preceding views.
Other methods~\cite{tseng2023consistent, photometricnvs} enhance novel view synthesis by employing cross-view attention or enforcing epipolar constraints within diffusion models.
MotionCtrl~\cite{wang2024motionctrl} leverages geometric priors from pre-trained video diffusion models to synthesize camera-controllable videos.
These methods are \textit{trajectory-based}, requiring input views to follow continuous spatial paths.
In contrast, \textit{image-set models} like CAT3D~\cite{gao2024cat3d} and more recently SEVA~\cite{seva} synthesize novel views from sparse and unordered input images, without trajectory constraints.
However, these methods are computationally expensive, with $O(n^2)$ complexity for attention, limiting conditioning views to small context windows containing only recently generated frames.
This limitation impacts long-term consistency of generated images.

Several concurrent works explore memory-based approaches for scene generation.
Genie~\cite{bruce2024genie, parker-holder24genie} maintains memory through recurrent state features.
Gen3C~\cite{ren2025gen3c} conditions on stored point clouds like ViewCrafter~\cite{yu2024viewcrafter}, inheriting inpainting-based limitations.
StarGen~\cite{zhai2025stargen} and WorldMem~\cite{xiao2025worldmemlongtermconsistentworld} use distance- and field-of-view-based spatial memory retrieval, which can recover spatially correlated views but lack geometric reasoning for occlusions.
In contrast, our surfel-based memory explicitly models occlusions and provides more principled view selection.

\section{Method}%
\label{sec:method}

Let $\mathbf{x}_1 \in \mathbb{R}^{H\times W\times 3}$ be an RGB image of a scene that we wish to explore, and let $\{\mathbf{c}_t\}_{t=1}^T$ be a sequence of camera parameters specifying a path through the scene.
Our goal is to generate a corresponding video, \ie, a sequence of images $\{\mathbf{x}_t\}_{t=1}^T$, where $\mathbf{x}_1$ is the given input image, and each subsequent frame $\mathbf{x}_t$ is consistent with the previous ones and reflects the specified cameras $\{\mathbf{c}_t\}_{t=1}^T$.
Moreover, we aim to do so autoregressively, generating $M$ novel views at a time for the given cameras (with $M$ kept small to ensure interactivity) while ensuring consistency with previously generated views.
This enables interactive scene exploration, where the user can specify the next camera positions at each step.

Specifically, we work with a sequence of \textit{views}, where each view $v_t = (\mathbf{x}_t, \mathbf{c}_t)$ consists of an image $\mathbf{x}_t$ and its corresponding camera $\mathbf{c}_t$ at timestep $t$.
At generation step $s$ ($s\geq 0$), we have generated $T = sM$ frames so far.
The goal is to generate $M$ new RGB images $\{\mathbf{x}_{T+m}\}_{m=1}^M$ for the next target camera parameters $\{\mathbf{c}_{T+m}\}_{m=1}^M$, given the previously generated views $\mathcal{V}^{(s)} = \{v_1, v_2, \dots, v_T\}$.
After generation, we update $T \leftarrow T + M$.

A challenge with this approach is that the context $\mathcal{V}^{(s)}$ grows unbounded over time.
Video generators addressing this problem~\cite{geogpt, photometricnvs, ren2022look, wang2024motionctrl} typically consider a fixed-length subset of $\mathcal{V}^{(s)}$, conditioning their generation only on the most recent $L$ views $\{v_{T-L+1}, \dots, v_T\}$.
This limitation results in severe inconsistencies when the generated sequence extends beyond the (small) context window.

We solve this problem by dynamically retrieving the most relevant subset of past views, denoted as $\mathcal{V}^* \subseteq \mathcal{V}^{(s)}$.
To achieve this, we introduce an efficient data structure, the \mnamel, that stores and retrieves past views based on their approximate 3D geometry.
We first describe the \mnamel~(\cref{sec:spatial_mem}) and then introduce the novel view generator~(\cref{sec:generator}).
An overview of the method is provided in \cref{fig:method}.

\subsection{Surfel-indexed view memory}%
\label{sec:spatial_mem}

Consider the problem of generating the next views $\{\mathbf{x}_{T+m}\}_{m=1}^M$ of the scene as seen from the cameras $\{\mathbf{c}_{T+m}\}_{m=1}^M$.
The generator must consider the previously generated views and ensure $\{\mathbf{x}_{T+m}\}_{m=1}^M$ are consistent with those, while generating new content as needed.
However, not all past views are equally relevant to the novel views $\{\mathbf{x}_{T+m}\}_{m=1}^M$.
We prioritize past views that have likely observed the largest portion of the scene currently being generated.
For example, if the current locale is surrounded by walls, parts of the scene behind those walls are likely less relevant for generating a new view of that locale.
Conversely, views that share a similar visible region with the target view can provide more useful information for generating its content.

\begin{figure*}[t!]
\centering
\includegraphics[width=0.98\textwidth]{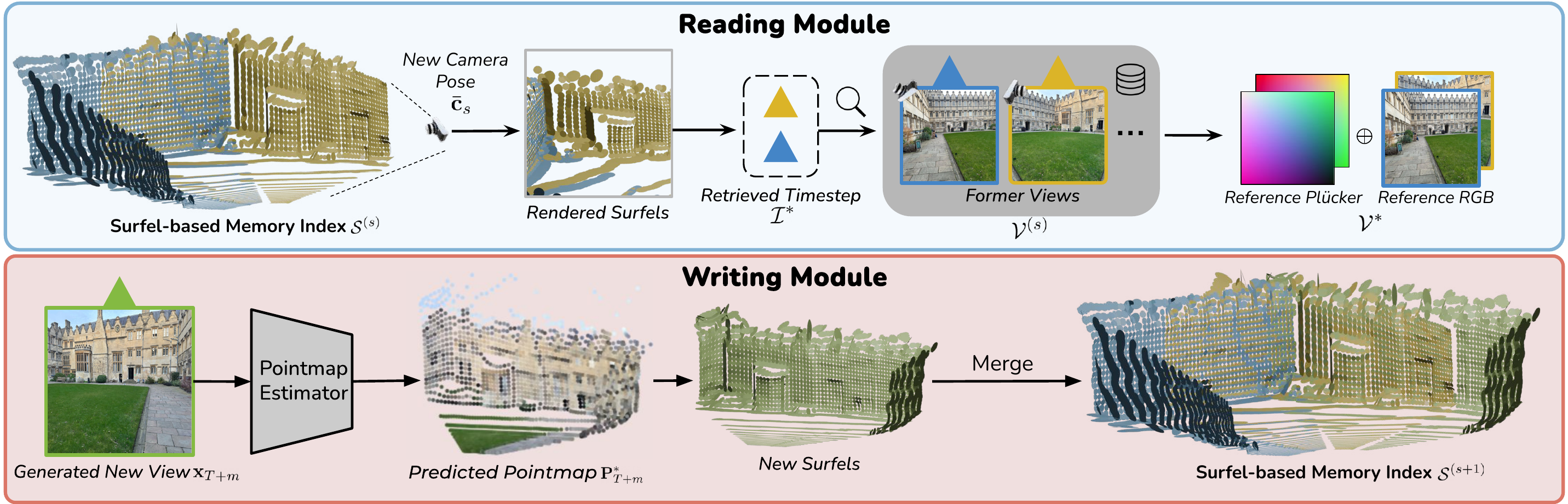}
\caption{\textbf{\mnamel. }
\emph{Reading} procedure renders surfels $\mathcal{S}^{(s)}$ with their attributes, containing past view indices as frame indices.  
We then select the $K$ most frequent frame indices in the rendered image to retrieve relevant past views from $\mathcal{V}^{(s)}$.  
\emph{Writing} procedure estimates geometry of newly generated views $\{\mathbf{x}_{T+m}\}_{m=1}^M$ as surfels and merges them with existing surfels.  
Frame indices $\{T+m\}_{m=1}^M$ are appended to surfels in these views, and novel views are stored, updating $\mathcal{V}^{(s)} \rightarrow \mathcal{V}^{(s+1)}$ and $\mathcal{S}^{(s)} \rightarrow \mathcal{S}^{(s+1)}$. 
}%
\label{fig:spatial_memory}
\end{figure*}

This principle drives our design for a view memory module.
We maintain a coarse model of the scene geometry using surfels (visualized in~\cref{fig:surfel_demo}), simple surface primitives that account for occlusion.
Each surfel stores the indices of the past views that observed it.
To generate a new view, the surfels visible from the new viewpoint are retrieved, using the associated indices to vote for which views to consider for generation.
This process is illustrated in \cref{fig:spatial_memory}.

Specifically, we define a \emph{surfel} as the tuple
\[
\mathbf{s}_k = \left(\mathbf{p}_k,\, \mathbf{n}_k,\, r_k,\, \mathcal{I}_k\right),
\]
where \(\mathbf{p}_k \in \mathbb{R}^3\) denotes the surfel's 3D position, \(\mathbf{n}_k \in \mathbb{R}^3\) is its surface normal, \(r_k \in \mathbb{R}\) is the surfel radius, and \(\mathcal{I}_k \subseteq \{1, 2, \ldots, T\} \) is a subset of past view indices that observed the surfel.
The surfel-based memory indexing of past views $\mathcal{V}^{(s)}$, encapsulating the scene, is represented as a set of surfels $\mathcal{S}^{(s)} = \{\mathbf{s}_k\}_{k=1}^{N^{(s)}}$, where $N^{(s)}$ is the number of surfels at generation step $s$.
We also maintain an octree to quickly retrieve surfels based on their geometry.
Next, we describe how to read from and write to this memory.

\paragraph{Reading from the memory.}
To retrieve the most relevant past views from the memory $\mathcal{V}^{(s)}$ for the novel cameras $\{\mathbf{c}_{T+m}\}_{m=1}^M$, we first compute their average pose $\bar{\mathbf{c}}_s \in \mathrm{SE}(3)$ as a reference (refer to~\acref{appen:average_camera}).
Then we render the surfels $\mathcal{S}^{(s)}$ from the averaged camera pose $\bar{\mathbf{c}}_s$.
Each surfel is rendered as a splat with its attributes, accounting for relative depth and occlusions, and contributing to the image based on its coverage.
The rendered attributes correspond to the indices of the past views.
The core intuition is that views observing the largest portion of the scene from the perspective of the novel camera are most relevant for novel view synthesis.
To identify these views, we rank past view indices based on their frequency across all rendered pixels.

We then select the top-$K$ most frequently represented indices $\mathcal{I}^*$ and use them to retrieve the corresponding past views, $\mathcal{V}^* = \{v_t\}_{t \in \mathcal{I}^*}$, from the memory $\mathcal{V}^{(s)}$.
This subset of views serves as the context for the new view generator, as described in \cref{sec:generator}.

To avoid oversampling repeatedly visited regions, we apply a non-maximum suppression algorithm that reduces redundancy in memory and promotes broader scene coverage among the top-$K$ views.
During retrieval, we retain only the most frequently referenced view among those with similar poses.
During writing to the memory, we merge surfels by comparing their associated camera poses—if two poses are highly similar, we discard the older one.

\paragraph{Writing to the memory.}
After generating the new views $\{v_{T+m}=(\mathbf{x}_{T+m},\mathbf{c}_{T+m})\}_{m=1}^M$, we add them to the memory $\mathcal{V}^{(s)}$ and update the surfel-based memory index $\mathcal{S}^{(s)}$.
We use an off-the-shelf point map estimator $\phi$, such as CUT3R~\cite{cut3r}, to estimate their point maps $\{\mathbf{P}^*_{T+m}\}_{m=1}^M$.
Specifically, we jointly estimate the point maps for the newly generated views along with the retrieved past views $\mathcal{V}^*$.
This joint estimation ensures that the new point maps are aligned with the coordinate frame of the existing scene geometry in the surfel-based memory index $\mathcal{S}^{(s)}$.

Next, we convert the estimated point maps to a set of new surfels.
For each newly generated frame $t \in \{T+1, T+2, \ldots, T+M\}$, we process its point map $\mathbf{P}^*_t$.
Since our approach only requires coarse geometry, we first downsample the point map $\mathbf{P}^*_t$ by a factor $\sigma$, yielding a smaller point map $\mathbf{P}_t \in \mathbb{R}^{H' \times W' \times 3}$, where $H' = H/\sigma$ and $W' = W/\sigma$.
For each pixel location $(u,v)$ in the downsampled point map, we have a 3D point $\mathbf{p}_{u,v,t}$, and we create a new surfel $\mathbf{s}_{k'}$ centered on it.
We compute each surfel's normal using the cross-product of displacement vectors to neighboring pixels.
For a pixel at 2D location $(u,v)$ in the image grid, we use neighboring pixels to estimate the local surface normal:
\[
\mathbf{n}_{k'} = \frac{(\mathbf{p}_{u+1,v,t} - \mathbf{p}_{u-1,v,t}) \times (\mathbf{p}_{u,v+1,t} - \mathbf{p}_{u,v-1,t})}{\left \| (\mathbf{p}_{u+1,v,t} - \mathbf{p}_{u-1,v,t}) \times (\mathbf{p}_{u,v+1,t} - \mathbf{p}_{u,v-1,t}) \right \|},
\]
where $\mathbf{p}_{u,v,t}$ represents the 3D point at image coordinates $(u,v)$ in frame $t$.
To ensure that the surfels reasonably cover the scene, we use a heuristic to compute the surfel's radius, which is proportional to the depth of the surfel and inversely proportional to the focal length and the cosine of the angle between the surfel normal and the viewing direction:
\[
r_{k'} = \frac{\frac{1}{2}\mathbf{D}_{u,v,t}/f_t}{\alpha + (1 - \alpha)\left \| \mathbf{n}_{k'} \cdot (\mathbf{p}_{u,v,t} - \mathbf{O}_t) \right \|},
\]
where $\mathbf{D}_{u,v,t}$ represents the depth at pixel $(u,v)$ in frame $t$, $\mathbf{O}_t$ is the camera center of frame $t$, $\mathbf{p}_{u,v,t} - \mathbf{O}_t$ is the viewing direction, and $f_t$ is the focal length at time $t$.
The factor $\alpha$ is used to avoid extreme values when $\left \| \mathbf{n}_{k'} \cdot (\mathbf{p}_{u,v,t} - \mathbf{O}_t) \right \| \approx 0$.

Finally, we check if the index $\mathcal{S}^{(s)}$ already contains a surfel $\mathbf{s}_k$ similar to the new surfel $\mathbf{s}_{k'}$.
Two surfels match if their centers are within a distance $d$ and the cosine similarity between their normals is above a threshold $\theta$.
If such a surfel $\mathbf{s}_k$ is found, we add the frame index $t$ to the surfel's set of past view indices: $\mathcal{I}_k \leftarrow \mathcal{I}_k \cup \{t\}$, and discard $\mathbf{s}_{k'}$.
If not, we set $\mathcal{I}_{k'} = \{t\}$ and add $\mathbf{s}_{k'}$ to the index $\mathcal{S}^{(s+1)}$.
This process transforms the surfel memory from $\mathcal{S}^{(s)}$ to $\mathcal{S}^{(s+1)}$ with $N^{(s+1)}$ surfels for the next generation step.

\subsection{Novel view generator}%
\label{sec:generator}

\begin{figure*}[ht!]
\centering
\includegraphics[width=0.95\textwidth]{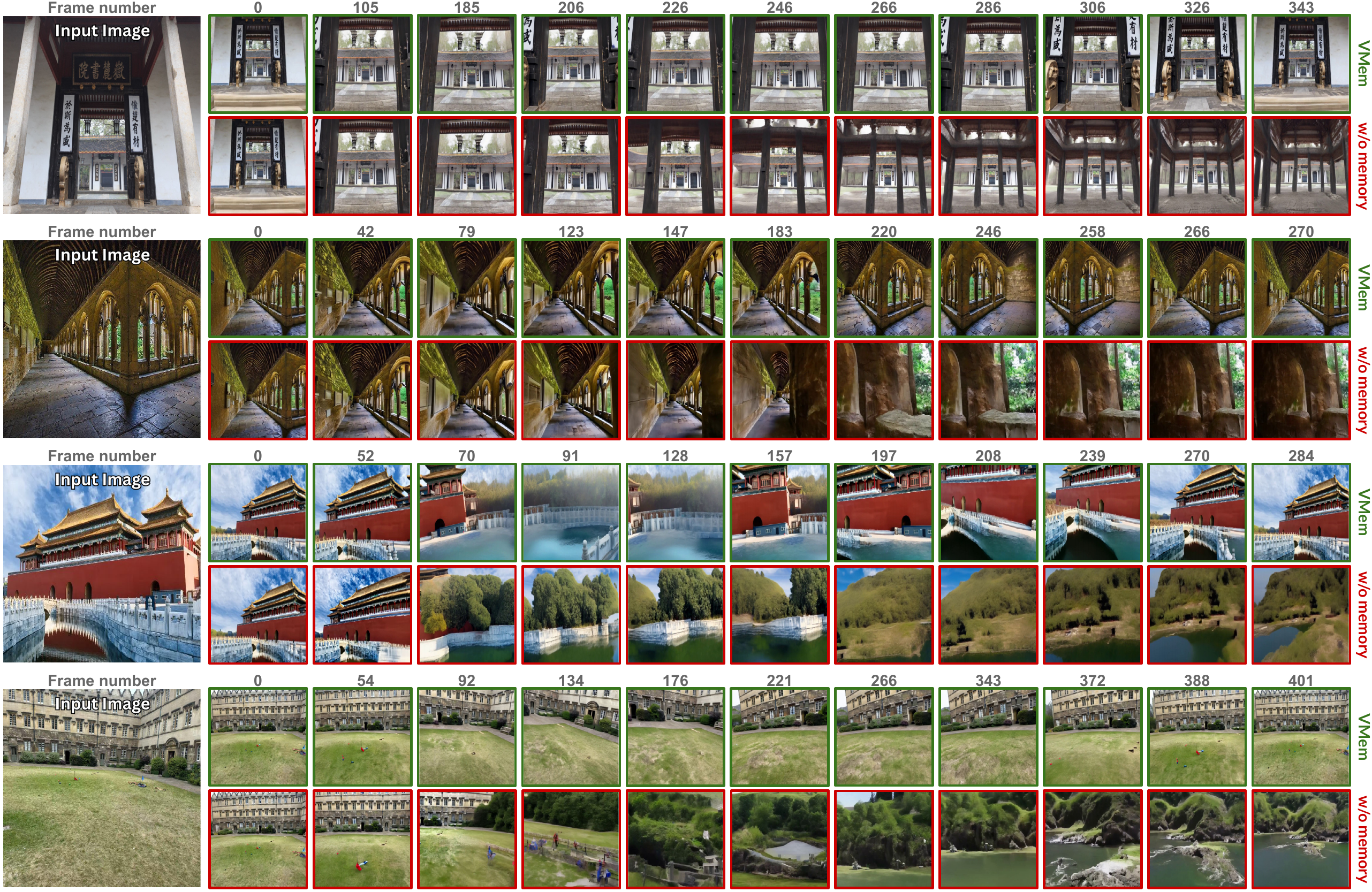}
\caption{\textbf{Long sequences with revisitations.}  
We compare our \mname~against a baseline without 
memory that relies solely on the last $K$ frames for context.
Each sequence: input images (left), then generated images at selected frames.  
Our method (\textcolor{OliveGreen}{top rows}) maintains consistency when revisiting observed regions, while the baseline (\textcolor{BrickRed}{bottom rows}) shows severe inconsistencies across extended sequences.}%
\label{fig:more_demo}
\end{figure*}

We use a camera-conditioned image-set generator $\psi$ to generate novel views.
The generator $\psi$ takes $K$ retrieved reference views $\mathcal{V}^* = \{v_t=(\mathbf{x}_t, \mathbf{c}_t)\}_{t \in \mathcal{I}^*}$ and the target camera poses $\{\mathbf{c}_{T+m}\}_{m=1}^M$ for the $M$ new frames to be generated and samples the novel views:
\[
\{\mathbf{x}_{T+m}\}_{m=1}^M \sim \psi\left( \{(\mathbf{x}_t, \mathbf{c}_t)\}_{t \in \mathcal{I}^*}, \{\mathbf{c}_{T+m}\}_{m=1}^M \right),
\]
where $\mathcal{I}^* \subseteq \{1, 2, \ldots, T\}$ are the frame indices of the retrieved reference views.

Specifically, we base our generator on the recent SEVA~\cite{seva} model.
Given the plug-and-play nature of our memory module, \mname~can work with other image-set generators as well.
We also fine-tune a more efficient version of SEVA that operates with a reduced number of reference frames.
We provide more details in \cref{sec:experimental_setup}.

\begin{table*}[t]
\centering
\small
\renewcommand{\arraystretch}{1.1}
\scalebox{0.85}{
\begin{tabular}{lcccccc|cccccc}
\toprule
\multicolumn{1}{c}{} &
\multicolumn{6}{c}{\textbf{Short-term} ($50^{th}\text{frame}$)} &
\multicolumn{6}{c}{\textbf{Long-term} ($\geq 200^{th}\text{frame}$)}\\
\cmidrule(lr){2-7}\cmidrule(lr){8-13}
\multicolumn{1}{c}{} &
\textbf{LPIPS} $\downarrow$ &
\textbf{PSNR} $\uparrow$ &
\textbf{SSIM} $\uparrow$ &
\textbf{FID} $\downarrow$ &
$R_{\text{dist}}$ $\downarrow$ &
$T_{\text{dist}}$ $\downarrow$ &
\textbf{LPIPS} $\downarrow$ &
\textbf{PSNR} $\uparrow$ &
\textbf{SSIM} $\uparrow$ &
\textbf{FID} $\downarrow$ &
$R_{\text{dist}}$ $\downarrow$ &
$T_{\text{dist}}$ $\downarrow$ \\
\midrule
   GeoGPT~\cite{geogpt}
    & 0.444 & 13.35  & --- & 26.72 & ---&---
    & 0.674 & 9.54  & --- & 41.87 & ---& ---\\
   Lookout~\cite{ren2022look}
    & 0.378 & 14.43  & --- &28.86  & 1.241& 0.876
    & 0.658 & 10.51  & --- &  58.12& 1.142&0.924\\
   PhotoNVS~\cite{photometricnvs}
    &  0.333& 15.51  & ---&21.76  & ---& ---
    & 0.588 &  11.54 & --- & 41.95 &--- &--- \\

   GenWarp~\cite{seo2024genwarp}
    & 0.436 &12.03   & 0.144& 29.69 & 0.564& 0.059 
    & 0.613 & 9.56  & 0.085 &  36.40& 0.850& \textbf{0.251} \\
   MotionCtrl~\cite{wang2024motionctrl}
    & 0.424  & 12.00   & 0.148 & 19.25 & 0.341 & 0.356
    & 0.605 & 9.13  & 0.083 & 35.45 & \textbf{0.748} & 0.609\\

   ViewCrafter~\cite{yu2024viewcrafter}
    & 0.377  & 16.97  & 0.262 & 25.39 &1.562 &0.208
    &  0.592 & 9.74  & 0.148 &  34.12& 2.177& 0.827\\

    SEVA~\cite{seva} ($K=17$)
    &  0.293 & 18.33  & 0.382  & 17.29 & 0.223& 0.118
    & 0.455 & 13.98  & 0.216 & 23.82 & 1.125 & 0.742\\

    \midrule

    \mname~($K=4$)
    & \textbf{0.287} & \textbf{18.49}  & \textbf{0.406} &\textbf{17.12}  & \textbf{0.219}& \textbf{0.039}
    & 0.493 & 13.12 & 0.183 &  27.15& 0.811& 0.499 \\

    \mname~($K=17$)
    & 0.293& 18.33  & 0.382 & 17.29 &0.223 & 0.118
    & \textbf{0.452}& \textbf{14.09} &  \textbf{0.227} & \textbf{23.56}  & 0.982 & 0.432 \\
\bottomrule
\end{tabular}
}
\caption{\textbf{Single-view NVS on RealEstate10K.}
Frames are subsampled with 10-frame intervals.
$K$ denotes the number of context views.
In the short-term setting, all past frames fit in the context windows for both SEVA~\cite{seva} and \mname; hence, SEVA~\cite{seva} and \mname~($K=17$) yield identical results since \mname~uses SEVA as the backbone.
\mname~($K=4$) is fine-tuned on RealEstate10K for this setting, yielding better performance than SEVA~\cite{seva} and \mname~($K=17$) in short-term scenarios.
\mname~surpasses baselines on most metrics in this benchmark. However, its \textbf{true advantage—spatial consistency—is not fully reflected} because RealEstate10K trajectories rarely revisit previously observed areas.
We address this limitation in~\cref{exp:cycle}.
}%
\label{tab:long-term}
\vspace{-1mm}
\end{table*}

\begin{table}[t]
\centering
\setlength{\tabcolsep}{3pt}
\scalebox{0.76}{
\begin{tabular}{lcccccc}
\toprule
\textbf{Method} & \textbf{LPIPS}~$\downarrow$ & \textbf{PSNR}~$\uparrow$  & \textbf{SSIM}~$\uparrow$& \textbf{FID}~$\downarrow$  & $R_{\text{dist}}$ $\downarrow$ & $T_{\text{dist}}$ $\downarrow$ \\
\midrule
Look-out~\cite{ren2022look}
&  0.809&   8.41&  0.069&  38.34&  1.262&0.341\\
GenWarp~\cite{seo2024genwarp}
&  0.507&   11.13&  0.134&  32.94&  0.848&0.241\\

MotionCtrl~\cite{wang2024motionctrl}
& 0.589 & 9.07  &  0.096& 26.86 &  0.871&0.293\\

ViewCrafter~\cite{yu2024viewcrafter}
&  0.401&   11.82&  0.217& 24.72 & 0.902&  0.492\\

SEVA~\cite{seva} ($K=17$)
&  0.401&   11.82&  0.217& 24.72 & 0.902&  0.492\\

\midrule

\mname~($K=4$)&  0.397&  15.72&  0.297& 24.97&  \textbf{0.821} & 0.392\\

\mname~($K=17$)&  \textbf{0.304}&  \textbf{18.15}&  \textbf{0.377}& \textbf{24.18}&  0.892& \textbf{0.165}\\

\bottomrule
\end{tabular}
}
\caption{\textbf{Results on cycle trajectories from RealEstate10K.} 
}%
\label{tab:cycle}
\end{table}

\begin{table}[t]
    \centering
    \setlength{\tabcolsep}{5pt}
    \scalebox{0.76}{
    \begin{tabular}{lccccc}
    \toprule
    \textbf{Method} & \textbf{LPIPS}~$\downarrow$ & \textbf{PSNR}~$\uparrow$ & \textbf{SSIM}~$\uparrow$  & $R_{\text{dist}}$ $\downarrow$ & $T_{\text{dist}}$ $\downarrow$ \\
    \midrule
   Look-out~\cite{ren2022look}
    &  0.792&   8.52&  0.008&  \textbf{0.727}&  1.499\\
   GenWarp~\cite{seo2024genwarp}
    &  0.521&   10.27&  \textbf{0.129}& 0.785&  1.982\\

   MotionCtrl~\cite{wang2024motionctrl}
    &  0.692&   8.21&  0.082& 0.842& \textbf{0.129}\\

   ViewCrafter~\cite{yu2024viewcrafter}
       &0.494& 13.62 & \textbf{0.129}  &1.643 & 0.492 \\

    \midrule

   \mname~($K=4$)
    & \textbf{0.472} & \textbf{14.11} & 0.121 & 1.204 &0.387 \\

    \bottomrule
    \end{tabular}
    }
    \caption{\textbf{Results on cycle trajectories from Tanks and Temples.} 
    }%
    \label{tab:tank}
    \vspace{0mm}
\end{table}

\section{Experiments}%
\label{sec:experiments}

We evaluate our method on established benchmarks for camera-conditioned autoregressive view generation from a single image, comparing our models to previous open-source approaches (\cref{exp:short-term}).
These benchmarks, however, have a critical limitation: camera trajectories rarely revisit previously observed regions, unlike real-world behavior where humans or robots often return to or look at previously explored areas.
To address this limitation, we propose a cycle-trajectory evaluation protocol that extends these benchmarks by making the camera return to the starting position along the same path in reverse order (\cref{exp:cycle}).
Additionally, we collect a small dataset with trajectories that revisit observed regions for qualitative evaluation in~\cref{fig:more_demo}.
Finally, we conduct an ablation study in \cref{exp:ablation}.

\subsection{Experimental setup}%
\label{sec:experimental_setup}

\paragraph{Implementation details.}%
We use pre-trained SEVA~\cite{seva} as the generator $\psi$.
SEVA uses a fixed total number of reference and target frames ($K+M=21$).
For our experiments, we use $M=4$ target views, which leaves $K=17$ context views.
However, this configuration demands substantial computational resources that may be prohibitive for many applications.
To address this limitation, we fine-tune a more efficient version with LoRA~\cite{hu22lora:} on the RealEstate10K training split~\cite{realestate} that works with a reduced number of reference frames ($K=4$) and target views ($M=4$), and use this efficient version for our main experiments.
We demonstrate (\cref{exp:cycle,exp:ablation}) that this fine-tuned version, when combined with~\mname, achieves comparable performance to the original SEVA while delivering approximately $12\times$ speedup at inference time.
We provide more technical details in~\acref{appen:implementation}.

\paragraph{Datasets.}%
\label{sec:datasets}
We consider two real-world datasets for novel view generation evaluation: RealEstate10K~\cite{realestate}, comprising indoor scene video clips, and Tanks-and-Temples~\cite{knapitsch2017tanks}, including indoor and outdoor scenes with larger camera motions.
For qualitative evaluation, we use in-the-wild images collected from the internet or captured with phone cameras as input views (see \cref{fig:more_demo}).

\paragraph{Evaluation metrics.}%
\label{sec:metrics}
We evaluate generated novel view quality based on three factors:
(1) overall quality of generated novel views, comparing generated novel view distributions to test set distributions using Fréchet Image Distance (FID)~\cite{fid};
(2) the ability of the model to preserve image details across views, measuring peak signal-to-noise ratio (PSNR) for pixel differences, along with LPIPS~\cite{lpips} and SSIM~\cite{ssim}, as in~\cite{ren2022look}; and
(3) alignment between generated camera poses and ground truth, following~\cite{he2024cameractrl}, where we express estimated camera poses relative to the first frame and normalize translation by the furthest frame.
We use DUSt3R~\cite{dust3r_cvpr24} to extract poses from generated views.
We calculate rotation distance $R_{\text{dist}}$ by comparing ground truth and extracted rotation matrices of each generated sequence:
\[
R_{\text{dist}} = \arccos\left(0.5\left(\text{tr}(\mathbf{R}_{\text{gen}} \mathbf{R}^{T}_{\text{gt}}) - 1\right)\right),
\]
where $\mathbf{R}$ denotes a rotation matrix.
We compute translation distance $t_{\text{dist}}$ as:
\[
t_{\text{dist}} = \left \| \mathbf{t}_{\text{gt}} - \mathbf{t}_{\text{gen}} \right \|_2.
\]

\begin{figure*}[ht]
\centering
\includegraphics[width=0.99\textwidth]{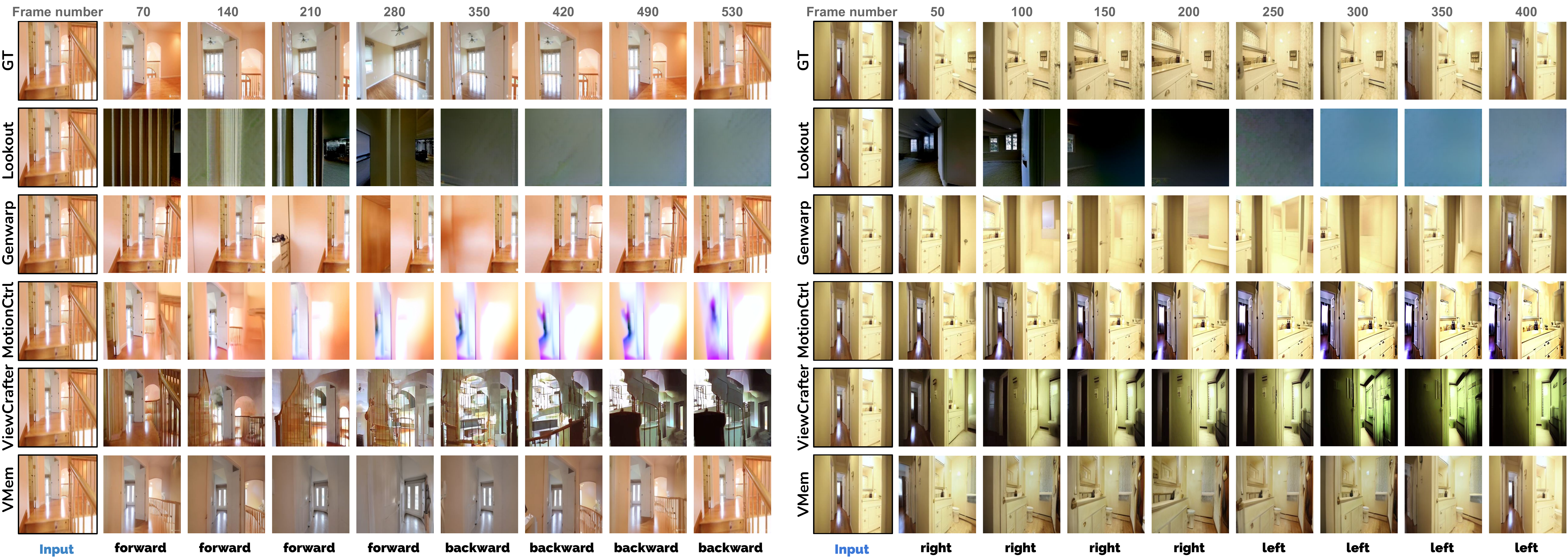}
\caption{\textbf{Qualitative comparison of the cycle trajectory frames} ($\geq 400$ frames). \mname~autoregressively generates new views while maintaining memory of all previously generated frames, ensuring consistency when revisiting previously seen locations.}%
\label{fig:longqualitative_comparison}
\end{figure*}

\subsection{Short- and long-term view generation}%
\label{exp:short-term}

We first evaluate \mname~on the benchmark from~\cite{ren2022look, photometricnvs}, which defines two scenarios: \emph{short-term} and \emph{long-term} novel view generation.
Beginning from the first frame of each ground-truth test sequence, the model autoregressively generates images along the ground truth camera trajectories with 10-frame intervals.
The \emph{short-term} scenario evaluates the fifth generated image (50 frames from the initial view), while the \emph{long-term} scenario evaluates the final image ($\geq$200 frames from the initial view).
Following the protocol, we only consider samples with more than 200 frames for the \emph{long-term} scenario and more than 50 frames for the \emph{short-term} scenario.
We report results in \cref{tab:long-term}.
\mname~substantially outperforms all baselines on the key metrics, demonstrating superior scene extrapolation.
However, the benchmark trajectories rarely revisit observed regions, limiting evaluation of long-term scene consistency.

\begin{table}[]
\centering
\setlength{\tabcolsep}{2.5pt}
\scalebox{0.682}{
\begin{tabular}{clccccc}
\toprule
\makecell{\textbf{Context} \\ \textbf{views $K$}}& \textbf{View Retrieval} & \textbf{LPIPS}~$\downarrow$ & \textbf{PSNR}~$\uparrow$ & \textbf{SSIM}~$\uparrow$ & $R_{\text{dist}}$~$\downarrow$ & $T_{\text{dist}}$~$\downarrow$ \\
\midrule
\multirow{4}{*}{\textbf{17}} 
& Temporal         & 0.477 & 13.92 & 0.188 & 0.976 & 0.254 \\
& Camera Distance (SEVA~\cite{seva})  & 0.397 & 15.72 & 0.297 & \textbf{0.821} & 0.392 \\
& Field of View & 0.374 & 15.75 & 0.292 & 0.911 & 0.382 \\
& \mname             & \textbf{0.304} & \textbf{18.15} & \textbf{0.377} & 0.892 & \textbf{0.165} \\
\midrule
\multirow{4}{*}{\textbf{4}} 
& Temporal         & 0.794 & 7.52  & 0.018 & 1.942 & 0.458 \\
& Camera Distance  & 0.422 & 13.27 & 0.187 & 1.787 & 0.319 \\
& Field of View & 0.424 & 13.11 & 0.192 & 1.782 & 0.285 \\
& \mname             & \textbf{0.381} & \textbf{14.82} & \textbf{0.275} & \textbf{0.793} & \textbf{0.124} \\
\bottomrule
\end{tabular}
}
\caption{\textbf{Ablation study on cycle trajectories from RealEstate10K.} We compare different view retrieval strategies for two context view counts $K$
}%
\label{tab:ablation}
\vspace{-4mm}
\end{table}

\subsection{Long-term view generation with revisitations}%
\label{exp:cycle}

To evaluate long-term scene consistency when revisiting observed regions—a scenario that rarely occurs in existing benchmarks—we propose a cycle-trajectory evaluation protocol as a proxy for real revisitation scenarios in human and robotic exploration.
In this protocol, the model generates frames following trajectories from initial to final poses, then returns along the same path to the starting position.
We evaluate metrics every 10 frames on the return trajectories, using all test sequences.
Following this protocol, we report quantitative results in \cref{tab:cycle} and qualitative comparisons in \cref{fig:longqualitative_comparison}.
\mname~outperforms all baselines across all metrics, demonstrating superior visual consistency when generating previously observed regions.

To validate generalization, we evaluate on Tanks-and-Temples containing indoor and outdoor scenes with more dynamic camera trajectories.
We use all six advanced scenes and apply the same \emph{cycle-trajectory} protocol using the first 50 frames, evaluating every frame without temporal subsampling.
Results are in \cref{tab:tank}.
\mname~consistently outperforms all baselines across in-domain (\cref{tab:cycle}) and out-of-domain (\cref{tab:tank}) evaluations on the majority of metrics.

We also qualitatively evaluate \mname~across diverse in-the-wild scenes in \cref{fig:more_demo}, demonstrating generation of diverse, consistent, high-quality long videos.

\subsection{Ablation study}%
\label{exp:ablation}

We ablate our surfel-based memory indexing for context view retrieval in \mname.
We evaluate our approach in the cycle-trajectory setting against three alternative retrieval strategies:
(1) temporal-based retrieval, selecting the most recent $K$ views from memory,
(2) camera distance-based retrieval, selecting top $K$ views with cameras closest to the target view camera—also used by SEVA~\cite{seva}, our backbone generator, and
(3) field-of-view-based retrieval, selecting top $K$ views with highest field-of-view overlap with the target view.
We also evaluate performance trade-offs between our lightweight, fine-tuned generator using $K=4$ context views versus original SEVA with $K=17$ context views.
We report results in \cref{tab:ablation}.

Our findings show \mname~consistently improves generator performance across both numbers of context views.
Improvement is most pronounced with fewer context views ($K=4$), highlighting the effectiveness of our surfel-based memory index.

\paragraph{Computational efficiency.}
A key practical advantage of our approach is its computational efficiency.
\mname~with only $K=4$ views achieves $12\times$ speed improvement over the original SEVA while recovering most of the performance achieved with $K=17$ views.
On an RTX 4090 GPU, our method generates frames in 4.2 seconds compared to 50 seconds for SEVA, representing a step towards real-time interactive scene exploration.

\section{Conclusion}%
\label{sec:conclusion}

We introduced \textit{\mnamel~(\mname)}, a novel plug-and-play memory module for long-term autoregressive scene generation using a surfel-indexed view memory.
By anchoring past views to a surfel-based representation of the scene and retrieving the most relevant ones, our approach improves long-term scene consistency while reducing computational costs.
Experiments on long-term scene synthesis benchmarks demonstrate that \mname~outperforms existing methods in scene coherence while enabling the use of fewer context views, leading to significantly faster generation.
Our work advances scalable video generation, with applications in virtual reality, gaming, and other domains.

\paragraph*{Acknowledgments.}

The authors of this work are supported by Clarendon scholarship, ERC 101001212-UNION, and AIMS EP/S024050/1. The authors would like to thank Xingyi Yang, Zeren Jiang, Junlin Han, Zhongrui Gui for their insightful feedback.

{%
\small%
\bibliographystyle{ieeenat_fullname}%
\bibliography{refs/vedaldi_general,refs/vedaldi_specific,refs/main}%
}

\newpage
\appendix
\begin{center}
\large\textbf{Appendix}
\end{center}

\section{Implementation details}
\label{appen:implementation}
To address the computational cost of SEVA~\cite{seva}, which uses a large fixed total number of reference and target frames ($K+M=21$), we fine-tune a more efficient version that employs a reduced number of reference frames ($K=4$) and target views ($M=4$).
We apply LoRA~\cite{hu22lora:} with rank $256$ and randomly sample context views online during training.
Training proceeds for 600,000 iterations on 8 A40 GPUs with a batch size of 24 per GPU, using the AdamW optimizer with a learning rate of $3\times 10^{-6}$, weight decay of $10^{-4}$, and a cosine annealing schedule.
For inference, we set the classifier-free guidance scale to $3$, the point map scaling factor $\sigma$ to $0.03$, and $\alpha$ to $0.2$ for surfel radius calculation.

\section{Average pose calculation}
\label{appen:average_camera}
To compute the average camera pose for rendering surfels, we average translations $\mathbf{t}_{T+m}$ with a simple mean, and rotations $\mathbf{R}_{T+m}$ by converting them to quaternions $\mathbf{q}_m$, aligning signs to a common hemisphere, and normalizing the mean quaternion:
\[
\bar{\mathbf{q}} = \frac{\sum_{m=1}^M \tilde{\mathbf{q}}_m}{\|\sum_{m=1}^M \tilde{\mathbf{q}}_m\|}, \quad \tilde{\mathbf{q}}_m = \text{sign}(\mathbf{q}_m \cdot \mathbf{q}_1) \cdot \mathbf{q}_m.
\]
The final average pose is $\bar{\mathbf{c}} = \begin{bmatrix} \mathrm{R}(\bar{\mathbf{q}}) & \bar{\mathbf{t}} \\ \mathbf{0}^\top & 1 \end{bmatrix}$, where $\mathrm{R}(\bar{\mathbf{q}})$ denotes the rotation matrix from $\bar{\mathbf{q}}$ and $\bar{\mathbf{t}} = \frac{1}{M}\sum_{m=1}^M \mathbf{t}_{T+m}$. 

\section{Autoregressive point map prediction}
Since we generate point maps for each view in an autoregressive manner, it is crucial to maintain their consistency across a shared coordinate space. Point-map estimators such as CUT3R include an optimization stage that jointly refines the depth, camera parameters, and point maps. To ensure a fixed camera trajectory, we freeze the camera parameters, which are user-defined inputs. Additionally, at each generation step when we have $T$ frames generated so far, we freeze all previously predicted depth maps for frames $1, 2, \ldots, T$ during optimization. This ensures that the resulting point maps and surfel representations remain consistent and causal. We then save the optimized depth maps of the newly generated frames $T+1, \ldots, T+M$ for future prediction.

\section{Limitations and discussion}

\paragraph{Evaluation protocol.}
Since there is no established benchmark for evaluating long-term consistency in scene video generation, we adopt cyclic trajectories as a proxy for assessment. However, these trajectories remain relatively simple and contain only limited occlusions, which means the full potential of \mname~in handling occlusions is not fully demonstrated. Moreover, existing evaluation metrics primarily capture low-level texture similarity in hallucinated content, rather than assessing true multi-view consistency—an inherent limitation of single-view autoregressive generation. As such, there is a clear need for more standardized evaluation protocols, which we leave for future exploration.

\paragraph{Limited training data and computing resources.} 
Due to limited computational resources, our more efficient version of the generator based on SEVA~\cite{seva} was fine-tuned only on the RealEstate10K dataset~\cite{realestate}. This dataset primarily consists of indoor scenes and a limited number of outdoor real-estate scenarios. Consequently, the model may struggle to generalize to broader contexts, with performance potentially degrading when dealing with natural landscapes or images containing moving objects compared to indoor environments. We believe this limitation stems primarily from insufficient dataset diversity rather than fundamental model constraints.

\paragraph{Inference speed.}
Due to the multi-step sampling process of diffusion models, \mname~requires 4.16 seconds to generate a single frame on an RTX 4090 GPU. This falls short of the real-time performance needed for applications such as virtual reality. We believe that future advancements in single-step image-set models and improvements in computational infrastructure hold promise for significantly accelerating inference speed.

\paragraph{Future improvements.}
Since our memory module relies heavily on the capabilities of the off-the-shelf image-set generator and the point map predictor, the performance of \mname~is expected to improve as these underlying models continue to advance.

\end{document}